\documentclass{article}
\usepackage{spconf,amsmath,epsfig}
\usepackage{booktabs} 
\usepackage{threeparttable}
\usepackage{epstopdf}
\usepackage{graphicx}
\usepackage{multicol}
\usepackage{multirow}
\usepackage[tight,footnotesize]{subfigure}
\usepackage{algorithm}
\usepackage{algpseudocode}
\usepackage{amsmath}
\usepackage{graphicx}
\usepackage{subfigure}   
\usepackage{cite}
 \usepackage{amsfonts,amssymb}

\title{Learning Binary Semantic Embedding for Histology Image Classification and Retrieval}
\name{Xiao Kang$^{1}$, Xingbo Liu$^{1*}$, Xiushan Nie$^{2}$,  Yilong Yin$^{1*}$ \thanks{Xingbo Liu and Yilong Yin are both the corresponding authors of this work.}}
\address{$^{1}$School of Software, Shandong University, Jinan, P.R. China\\
$^{2}$School of Computer Science and Technology, Shandong Jianzhu University, Jinan, P.R. China\\
\{sckx,sclxb\}@mail.sdu.edu.cn, niexsh@hotmail.com,  ylyin@sdu.edu.cn}

\begin{document}
\maketitle
\begin{abstract}
With the development of medical imaging technology and machine learning, computer-assisted diagnosis which can provide impressive reference to pathologists, attracts extensive research interests.  The exponential growth of medical images and uninterpretability of traditional classification models have hindered the applications of computer-assisted diagnosis. To address these issues, we propose a novel method for \underline{L}earning \underline{B}inary \underline{S}emantic \underline{E}mbedding (LBSE). Based on the efficient and effective embedding, classification and retrieval are performed to provide interpretable computer-assisted diagnosis for histology images. Furthermore, double supervision,  bit uncorrelation and balance constraint, asymmetric strategy and discrete optimization are seamlessly integrated  in the proposed method for learning binary embedding. Experiments conducted on three benchmark datasets validate the superiority of LBSE under various  scenarios.
\end{abstract}

\begin{keywords}
Computer-assisted diagnosis, Breast cancer, Histology image, Binary embedding, Approximate nearest neighbor search
\end{keywords}

\section{Introduction}
Although modern medicine makes progress with each passing day, the breast cancer still remains the second leading cause of death for women worldwide. Early detection and diagnosis are of much concern to prevent the women form death since the nosogenesis of this disease remains vague. The earlier to detect and diagnose, the easier to increase the success of treatment, save lives and bring down costs in healthcare. With the development of digital imaging technology, higher and higher percentage of medical diagnosis is taken up by histology image. Nevertheless, medical image analysis is rather labor-intensive and specialized, making the pathologists tired of strenuous effort. Fortunately, by the development of machine learning and medical imaging technology, computer-assisted diagnosis which can provide considerable assistance for pathologists, has become a hot topic in the field of computer vision and medical image processing \cite{duc2019classification, huang2020combining}. 

Existing methods for computer-assisted diagnosis based on histology image typically learn classification models (usually are black box) to predict the categories of given medical images. However, the ultimate prediction results are usually hard to explain. For example, a seasoned pathologist may diagnose histology images not only by their properties, but also through comparing them with previous cases. Considering that, in this study, we propose to adopt k-nearest neighbor classification by designing an effective retrieval method. In addition, with the advent of exponential growth in medical images, nearest neighbor search, which returns the most similar images in a large database, has been intractable in both time and storage. Therefore, the approximate nearest neighbor search within a large database has become momentous in this case. Hashing can encode image data into a binary embedding with similarity-preserving, which provides remarkable efficiency in both computational burden and memory requirements, and has been much compelling in approximate nearest neighbor search \cite{wang2018survey}.  Learning-based hashing \cite{ nie2016comprehensive,  nie2019joint,  nie2018robust, liu2018fast, liu2018modality, shen2015supervised, liu2016natural, gui2018fast, luo2018scalable, luo2018fast, gui2018r, liu2019SSLH,liu2019SDHMLR,liu2019moboost, chen2020strongly,liu2020model} is the well-researched hashing method that can provide superior retrieval performance through analyzing the latent characteristics of data. 

To provide interpretable diagnosis with both efficacy and efficiency, we propose a novel  method for  \underline{L}earning \underline{B}inary \underline{S}emantic \underline{E}mbedding (LBSE). Classification and retrieval tasks conducted on the binary semantic embedding are supposed to find the cases which are similar to the unseen inputs.  The main contributions of this work are three-fold:

\begin{itemize}
\item To the best of our knowledge, this study is the first endeavor that focuses on learning binary semantic embedding for the computer-assisted diagnosis of breast cancer based on histology images.

\item  We develop a supervised method for learning binary semantic embedding. In the proposed method, double supervision, bit uncorrelation and balance constraint, asymmetric strategy and discrete optimization, are seamlessly integrated to achieve better performance.

\item  Experiments on three widely-used datasets demonstrate the superiority of the proposed method under various scenarios.
\end{itemize}

\section{The Proposed Method}
In this section, we first introduce the  notations used in this study, then further elaborate the proposed method form three aspects, {\emph{i.e.}} binary code learning, optimization process and out-of-sample extension strategy.

\subsection{Notation}
Assume there is a training set consisting of $N$ instances, {\emph{i.e.}}, $\{ {{\bf{x}}_i}, {{\bf{y}}_i} \} _{i = 1}^N$ with ${{\bf{x}}_i}$  being $D$-dimensional feature vector of the $i_{th}$ instance. ${{\bf{y}}_i= \{ {{{y}}_{ic}}} \} \in \{0, 1\}^{C}$ is the label vector of the $i_{th}$ instance, where $C$ is the number of categories, and ${y_{ic}} = 1$ if ${{\bf{x}}_i}$ belongs to $c_{th}$ class, $c=1,\cdots, C$ and 0 otherwise. We utilize $\bf{X} \in \mathbb{R}^{D \times N}$ and $\bf{Y} \in \mathbb{R}^{C \times N}$ to represent the feature matrix and label matrix, respectively.

\subsection{Binary Code Learning}
Linear models are simple, yet effective ways to describe the relationships between semantic label  and hash code \cite{shen2015supervised, liu2016natural, gui2018fast, luo2018scalable, luo2018fast, gui2018r, liu2019SSLH,liu2019SDHMLR, chen2020strongly}. In this study, an orthogonal linear projection ${\bf{W}}$ is utilized to regress the label matrix ${\bf{Y}}$ to hash matrix ${\bf{H}}$ to stabilize the regression \cite{gui2018fast} and relieve the hubness problem  \cite{LazaridouHubness}. This process can  be formulated  as
\begin{equation}
\label{Loss_YH}
\begin{split}
&\min\limits_{{\bf {W}},{\bf {H}}}\left \| {\bf {H}}-{\bf{W}}^{T}\bf{Y} \right \|^{2},\\
& \textup{s.t.} \quad {\bf {W}}{\bf {W}}^{T} = {\bf{I}}, {\bf {H}}\in \{-1,+1\}^{L\times N},\\
\end{split}
\end{equation}
where $L$ is the length of hash code. 

The properties of the binary code, {\emph{a.k.a}} bit balance and bit uncorrelation, have been proved to be very significant \cite{gong2013iterative}. 
Bit balance means that each bit has the same chance of being $+1$ or $-1$, while bit uncorrelation means that any two bits are independent. We can formulate these two constraints as
\begin{equation}
{\bf{H1}} = {\bf{0}},\quad {\bf{H}}{{\bf{H}}^T} = N{\bf{I}},
\end{equation}
where $\bf{1}$ is an $N$-dimensional all-ones vector and $\bf{I}$ is an identity matrix of size $L$.

Nevertheless, it is much are intractable to utilize these two constraints directly owing to the binary quadratic problem \cite{kang2016column}. In this study, an auxiliary variable ${\bf{B}}$ is introduced to integrate these constraints skillfully. Specifically, to capture more relations among samples and generate more similarity-preserving embedding, pairwise similarity matrix ${\bf{S}}$ is embedded into the binary space by an asymmetric strategy. This process can be formulated as
\begin{equation}
\label{Loss_HB}
\begin{split}
&\min\limits_{{\bf {H}},{\bf {B}}}\left\|{\bf {H}}^{T}{\bf {B}}-L\cdot {\bf {S}}\right \|^{2}
+ \beta \left \| {\bf {H}}-\bf{B} \right \|^{2}\\
& \textup{s.t.} \quad {\bf{B}}{\bf {1}} = 0, {\bf {B}}{\bf {B}}^{T} = N{\bf{I}}, {\bf {H}}\in \{-1,+1\}^{L\times N},\\
 \end{split}
\end{equation}
where  $\beta$ is the hyperparameter. The utilization of semantic label and pairwise semantic similarity, {\emph{double supervision}}, can preserve more semantic information in the binary embedding \cite{luo2018scalable, luo2018fast}.

In addition, a projection $\bf{P}$ with $\ell_{2}$-norm regularization is used to describe the regression form feature matrix to hash matrix. The process can be written as 
\begin{equation}
\label{Loss_P}
\min\limits_{{\bf{P}}}\left\|{\bf{H}}-{\bf{P}}^{T}{\bf{X}} \right\|^{2}+\lambda \left\|{\bf{P}}\right\|^{2},
\end{equation}
where $\lambda$ is a hyperparameter. The $\ell_{2}$-norm regularization can smooth the solution, prevent over-fitting and improve the stability of linear regression \cite{hoerl1970ridge}.

Combining Eq.(\ref{Loss_YH}), Eq.(\ref{Loss_HB}) and Eq.(\ref{Loss_P}), we have the following final objective function  of the proposed method,
\begin{equation}
\label{Loss}
\begin{split}
&\min\limits_{{\bf {H}},{\bf {B}},{\bf {W}},{\bf {P}}}
\left\|{\bf {H}}^{T}{\bf {B}}-L\cdot {\bf {S}}\right \|^{2}+\alpha \left \| {\bf {H}}-{\bf{W}}^{T}\bf{Y} \right \|^{2}\\
&+  \beta \left \| {\bf {H}}-\bf{B} \right \|^{2}+\gamma \left \| {\bf {H}}-{\bf{P}}^{T}\bf{V} \right \|^{2}+\lambda \left \| {\bf{P}} \right \|^{2}\\
& \textup{s.t.} \quad {\bf {W}}{\bf {W}}^{T} = {\bf{I}}, {\bf{B}}{\bf {1}} = 0, {\bf {B}}{\bf {B}}^{T} = N{\bf{I}}, {\bf {H}}\in \{-1,+1\}^{L\times N},\\
 \end{split}
\end{equation}
where $\alpha$, $\beta$, $\gamma$ and $\lambda$ are hyperparameters.

\subsection{Optimization}
The problem in Eq. (\ref{Loss}) is noncontinuous and nonconvex, making it intractable to be optimized directly. In this study, we try to solve this challenging problem with following steps until convergence or a fixed number of iterations.

{\bf {W-Step}}: Learn the projection, $\bf{W}$, with the other variables unchanged. The problem in Eq. (\ref{Loss}) becomes
\begin{equation}
\label{solution-W1}
\min\limits_{{\bf {W}}}\left \| {\bf {H}}-{\bf{W}}^{T}\bf{Y} \right \|^{2}, \quad \textup{s.t.} \quad {\bf {W}}{\bf {W}}^{T} = {\bf{I}}.
\end{equation}
Since $ \left\| {\bf{W}}^{T}\bf{Y} \right \|^{2}=Tr({\bf{Y}}^{T}\bf{W}{\bf{W}}^{T}\bf{Y})=Tr({\bf{Y}}^{T}\bf{Y})$, Eq. (\ref{solution-W1}) can be rewritten as
\begin{equation}
\label{solution-W2}
\max\limits_{{\bf {W}}}
Tr({\bf{W}}({\bf{H}}{\bf {Y}}^{T})).
\end{equation}
The Procrustes problem in Eq. \ref{solution-W2} can be solved with analytic solutions \cite{Xia2015Sparse}. We first  perform
SVD ${\bf{H}}{\bf {Y}}^{T} = {\bf{U}} \sum {\bf{V}}^{T}$, where ${\bf{U}}$ is an $L \times L$  orthogonal matrix, $\sum$ is a $L \times C$ matrix and ${\bf{V}}$ is an $C \times C$ orthogonal matrix. 
Then the solution for ${\bf{W}}$ is 
\begin{equation}
\label{solution-W}
{\bf{W} = {\bf{V}}\hat{{\bf{U}}}}^{T},
\end{equation}
where $\hat{{\bf{U}}}$ contains first $C$ columns of ${\bf{U}}$.

{\bf {B-Step}}: Learn the auxiliary representation, $\bf{B}$, holding the other variables unchanged. The problem in Eq. (\ref{Loss}) becomes
\begin{equation}
\label{solution-B1}
\begin{split}
\min\limits_{{\bf {B}}}
&\left \|{\bf{H}}^{T}{\bf {B}}-L \cdot{\bf {S}}\right \|^{2}+ \beta \left \|{\bf {H}}-{\bf {B}} \right \|^{2}, \\ &\textup{s.t.} \quad {\bf {B}}{\bf {1}} = {\bf{0}}, {\bf {B}}{\bf {B}}^{T} = N{\bf{I}}.
\end{split}
\end{equation}

Since $\left \|{\bf{H}}^{T}{\bf {B}}\right \|^{2}=Tr({\bf{H}}^{T}{\bf {H}})=L*N$, the problem in Eq. (\ref{solution-B1}) can be rewritten as 
\begin{equation}
\label{solution-B3}
\begin{split}
&\max\limits_{{\bf {B}}} Tr({\bf{B}}^{T}({{\bf{H}}{\bf{S}}+\beta \bf{H}})), \\
&\textup{s.t.} \quad {\bf {B}}{\bf {1}} = {\bf{0}}, {\bf {B}}{\bf {B}}^{T} = N{\bf{I}}.
\end{split}
\end{equation}
For simplicity, we define $\bf{Q}= {\bf{H}}{\bf{S}}+ \beta{\bf{H}}$, ${\bf{J}}={\bf{I}}_{N}-\frac{1}{N}{\bf{1}}_{N}{\bf{1}}_{N}^{T}$.
Then, we perform the eigendecomposition for ${\bf{QJQ}}^{T}$ as
\begin{equation}
\label{solution-B4}
{\bf{QJQ}}^{T}=\left [ {\bf{Z}} \quad \ddot{{\bf{Z}}}\right ] \begin{bmatrix}
{\bf{\Omega}} & {\bf{ 0}}\\ 
{\bf{0}}&  {\bf{0}}
\end{bmatrix}\left [ {\bf{Z}} \quad \ddot{{\bf{Z}}}\right ]^{T},
\end{equation}
where ${\bf{\Omega}} \in {\mathbb{R}}^{{L'}\times{L'}}$ and ${\bf{Z}} \in \mathbb{R}^{{L'}\times{L'}}$ are the diagonal matrices of the positive eigenvalues and the corresponding eigenvectors, respectively. $\ddot{{\bf{Z}}}$ is the matrix of the remaining ${L}-{L'}$ eigenvectors corresponding to zero eigenvalue. ${L'}$ is the
rank of ${\bf{QJQ}}^{T}$. With a Gram-Schmidt process on $\ddot{{\bf{Z}}}$ , we can easily get an orthogonal matrix ${\tilde{{\bf{Z}}}} \in  \mathbb{R}^{{L}\times{(L-L')}} $ . We further define ${\bf{M}}={\bf{JQ}}^{T}{\bf{Z}}{\bf{\Omega}}^{-{\frac{1}{2}}}$, and a random orthogonal matrix ${\tilde{{\bf{M}}}} \in  {\mathbb{R}}^{{N}\times{(L-L')}}$. If ${L'}={L}$, $\ddot{{\bf{Z}}}$, $\tilde{{\bf{Z}}}$ and $\tilde{{\bf{M}}}$ are empty.
Finally, according to \cite{liu2014discrete}, we can obtain the solution of
${\bf{B}}$ as
\begin{equation}
\label{solution-B}
{\bf{B}}=\sqrt{N}\left [ {\bf{Z}} \quad \tilde{{\bf{Z}}}\right ] \left [ {\bf{M}} \quad \tilde{{\bf{M}}}\right ]^{T} 
\end{equation}

{\bf {H-Step}}: Learn the hash matrix, $\bf{H}$, while the other variables are fixed. The problem in Eq. (\ref{Loss})  becomes
\begin{equation}
\label{solution-H1}
\begin{split}
&\min\limits_{{\bf {H}}}
\left\|{\bf {H}}^{T}{\bf {B}}-L\cdot {\bf {S}}\right \|^{2}+\alpha \left \| {\bf {H}}-{\bf{W}}^{T}\bf{Y} \right \|^{2}\\
&+ \beta \left \| {\bf {H}}-\bf{B} \right \|^{2} +\gamma \left \| {\bf {H}}-{\bf{P}}^{T}\bf{V} \right \|^{2}\\
& \textup{s.t.} \quad  {\bf {H}}\in \{-1,+1\}^{L\times N},\\
 \end{split}
\end{equation}
Since $\left \|{\bf{H}}^{T}{\bf {B}}\right \|^{2}=Tr({\bf{H}}^{T}{\bf {H}})=L*N$, the problem in Eq. (\ref{solution-H1}) can be rewritten as 
\begin{equation}
\label{solution-H3}
\begin{split}
&\max\limits_{{\bf {H}}} Tr({\bf{H}}^{T}({\bf{B}}{\bf{S}}+\alpha{\bf{W}}^{T}{\bf{Y}}+\beta{\bf{B}}+\gamma {\bf{P}}^{T}\bf{V})), \\
 &\textup{s.t.} \quad {\bf {H}}\in \{-1,+1\}^{L\times N}.
\end{split}
\end{equation}
The analytic solution of ${\bf{H}}$ can be calculated discretely as
\begin{equation}
\label{solution-H}
    {\bf{H}}=sgn({\bf{B}}{\bf{S}}+ \alpha {\bf{W}}^{T}{\bf{Y}}+\beta{\bf{B}}+\gamma {\bf{P}}^{T}\bf{V}),
\end{equation}
where $sgn(\cdot)$ is a sign function. This process is done without relaxation, thus avoiding accumulated quantization error.

{\bf {P-Step}}: Learn the projection, ${\bf{P}}$, while holding the other variables fixed. The problem in Eq. (\ref{Loss}) becomes
 \begin{equation}
\min\limits_{{\bf{P}}}\left\|{\bf{H}}-{\bf{P}}^{T}{\bf{X}} \right\|^{2}+\lambda \left\|{\bf{P}}\right\|^{2}.
\end{equation}
Then, the closed-form solution of ${\bf{P}}$ is
 \begin{equation}
  \label{solution-P}
{\bf{P}}=({\bf{XX}}^{T}+\lambda {\bf{I}})^{-1}{\bf{X}}{\bf{H}}^{T}.
 \end{equation}
In conclusion, we try to solve the  nonconvex mixed integer optimization problem with an iterative framework based on the above four steps. 

\begin{table*}[htp]  
  \centering  
  \fontsize{9}{11}\selectfont  
  \begin{threeparttable}  
  \caption{Overall comparison of four evaluation matrices on the three benchmark datasets. The OA, Sen, PPV and F1 mean the Overall Accuracy, Sensitivity, Positive Predictive Value and F1-Score, respectively. The best results are shown in bold. }  
  \label{performance}  
    \begin{tabular}{c|cccc|cccc|cccc}  
    \toprule  
    \multirow {2}{*}{Method} &\multicolumn{4}{c}{BreaKHis 100X} &\multicolumn{4}{|c}{BreaKHis 200X} &\multicolumn{4}{|c}{BreaKHis 400X}\cr
    \cmidrule(lr){2-5} \cmidrule(lr){6-9} \cmidrule(lr){10-13} 
    &OA&Sen&PPV&F1 &OA&Sen&PPV&F1 &OA&Sen&PPV&F1\cr  
    \midrule  
SDH&0.3610&0.4174&0.5806&0.4846&0.6199&0.6383&0.7763&0.7000&0.4822&0.5069&0.6818&0.5809\cr
NSH&0.2831&0.3128&0.5000&0.3849&0.4730&0.5210&0.6493&0.5781&0.3600&0.4403&0.5600&0.4930\cr
FSDH&0.3588&0.4318&0.5702&0.4913&0.6008&0.6192&0.7607&0.6826&0.4960&0.5308&0.6849&0.5981\cr
R2SDH&0.3566&0.4277&0.5722&0.4885&0.6423&0.6527&0.7958&0.7169&0.5040&0.5233&0.6998&0.5987\cr
SSDH&0.7033&0.7464&0.8253&0.7833&0.6716&0.7695&0.7857&0.7767&0.4924&0.6428&0.6438&0.6422\cr
FSSH&0.8256&0.8029&0.9457&0.8683&0.7442&0.7900&0.8334&0.8106&0.5433&0.6221&0.7010&0.6590\cr
SDHMLR&0.8426&0.8226&0.8558&0.8798&0.8340&0.8383&0.8934&0.8074&0.6924&0.7157&0.7881&0.7895\cr
SCDH&0.8492&0.8409&0.9425&0.8888&0.8680&0.8563&0.9483&0.8999&0.6916&0.7384&0.8088&0.7718\cr
\hline
LBSE &\bf 0.8566&\bf 0.8431&\bf 0.9514&\bf 0.8939&\bf 0.8797&\bf 0.8671&\bf 0.9552& \bf  0.9090&\bf 0.8124&\bf 0.8201&\bf 0.9057&\bf 0.8608\cr
    \bottomrule  
    \end{tabular}  
   \end{threeparttable}  
\end{table*} 


\subsection{Out-of-sample Extension} 
For out-of-sample extension, the proposed LBSE utilizes the learned linear projection $\bf{P}$ to get the hash codes of unseen samples. Specifically, given a new query, we first perform feature extraction to get the corresponding feature vector ${\bf{a}}$. Then, the binary codes ${\bf{h}}$ for the new query can be obtained by ${\bf{h}}=sgn({\bf{P}}^{T}\varphi({\bf{a}}))$, where $sgn(z)=1$ if $z>0$ and $sgn(z)=-1$, otherwise. Based on the binary code of query, we can return the instances with the small Hamming distance in the retrieval database. Furthermore, we can predict the label of the query by K-nearest neighbor classification. It is noted that the K-nearest neighbor search is effective and efficient since the retrieval space is binary semantic embedding.

\section{Experiment}

\subsection{Experimental Settings}
To confirm the superiority of our method, we conduct sufficient experiments on three benchmark datasets, BreaKHis 100X, BreaKHis 200X and BreaKHis 400X \cite{fabio2016BreaKHis}. 

We compared the proposed LBSE with the supervised methods:
Supervised Discrete Hashing (SDH) \cite{shen2015supervised},
Natural Supervised  Hashing (NSH) \cite{liu2016natural},
Fast Supervised Discrete Hashing (FSDH) \cite{gui2018fast},
Scalable Supervised Discrete Hashing (SSDH) \cite{luo2018scalable},
Fast Scalable Supervised  Hashing (FSSH) \cite{luo2018fast},
Robust Rotated Supervised Discrete Hashing (R2SDH)  \cite{gui2018r},
Supervised Short-Length Hashing (SSLH) \cite{liu2019SSLH},
Supervised Discrete Hashing with Mutual Linear Regression (SDHMLR) \cite{liu2019SDHMLR},
and Strongly Constrained Discrete Hashing (SCDH) \cite{chen2020strongly}. 
Overall accuracy, sensitivity, positive predictive value and F1-score \cite{heng2010Criterion}  are adopted to evaluate the classification performance, while 
mean average precision and precision@K are utilized to evaluate the retrieval performance \cite{gui2018r, liu2019SSLH, liu2019SDHMLR}.

For fair comparison, a pretrained ResNet50 \cite{He2016ResNet} is utilized to perform feature learning for all of the baselines and the proposed method.
To verify the stability of the proposed method, we perform five runs for all of the baselines and the proposed method and average their performance for comparison. For the experimental parameters, we set $\alpha=0.5$, $\beta=5$, $\gamma=\lambda=10^{-5}$ by grid search.  In addition, we only consider top 99 retrieved samples. All experiments are conducted on a computer with an Intel Core i7-6700 3.40 GHz 4 processor and 32 GB RAM. And the programming environment is MATLAB R2019b.
The code for the proposed LBSE is released at {\emph{https://github.com/bd622/DiscretHashing.}}

\begin{figure}[htb]
\centering
\subfigure[BreaKHis 100X]{
\includegraphics[width=0.15\textwidth]{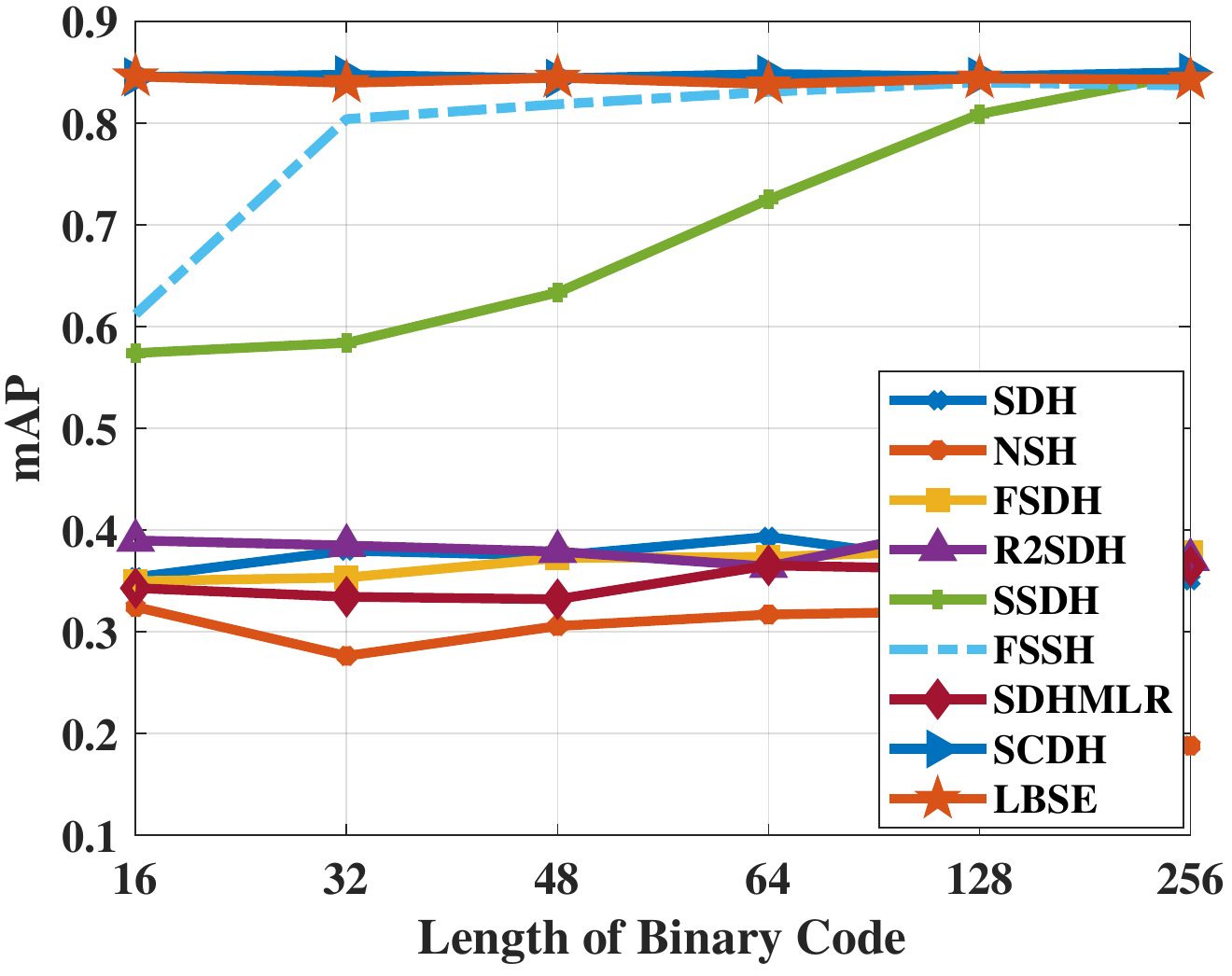}}
\subfigure[BreaKHis 200X]{
\includegraphics[width=0.15\textwidth]{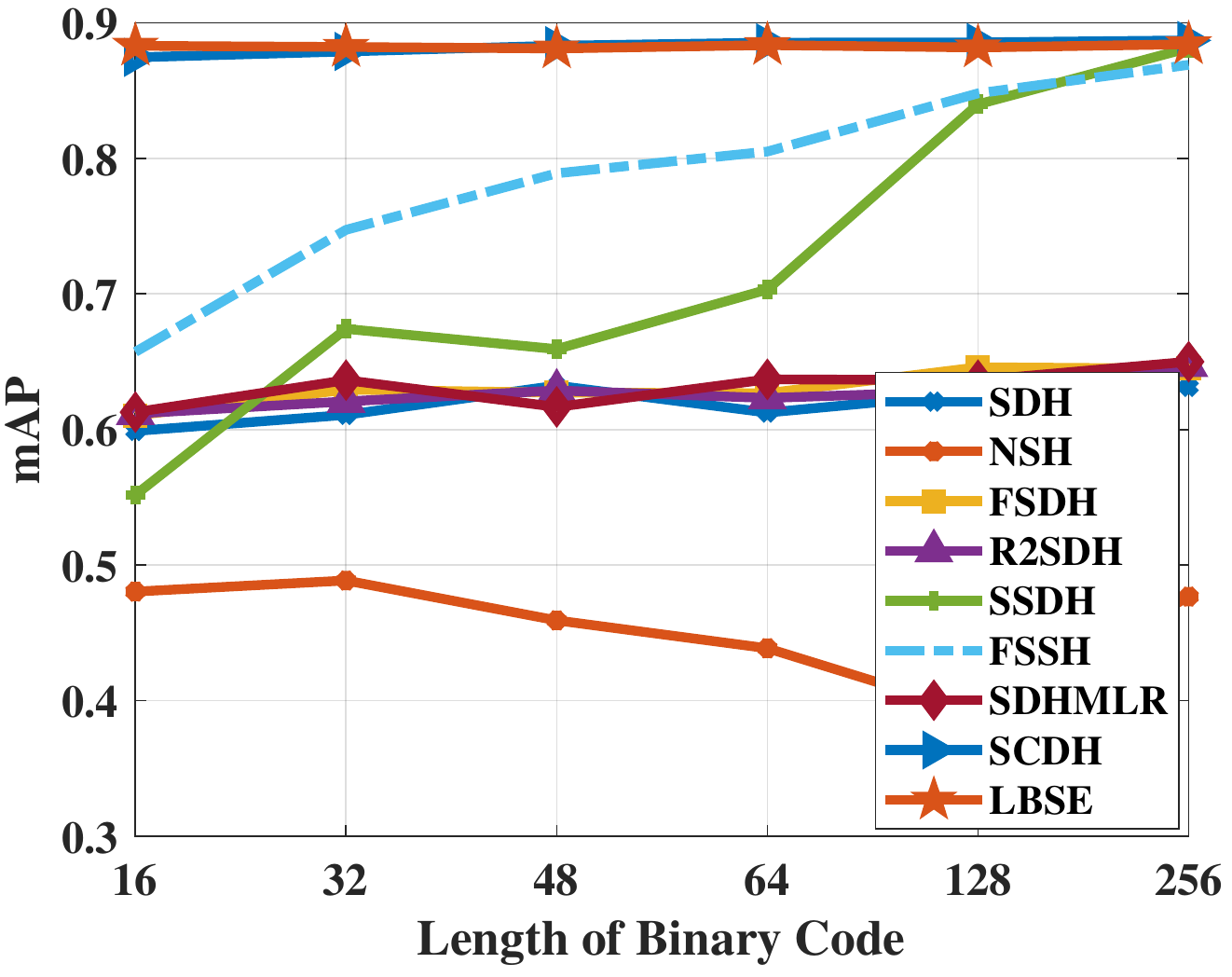}}
\subfigure[BreaKHis 400X]{
\includegraphics[width=0.15\textwidth]{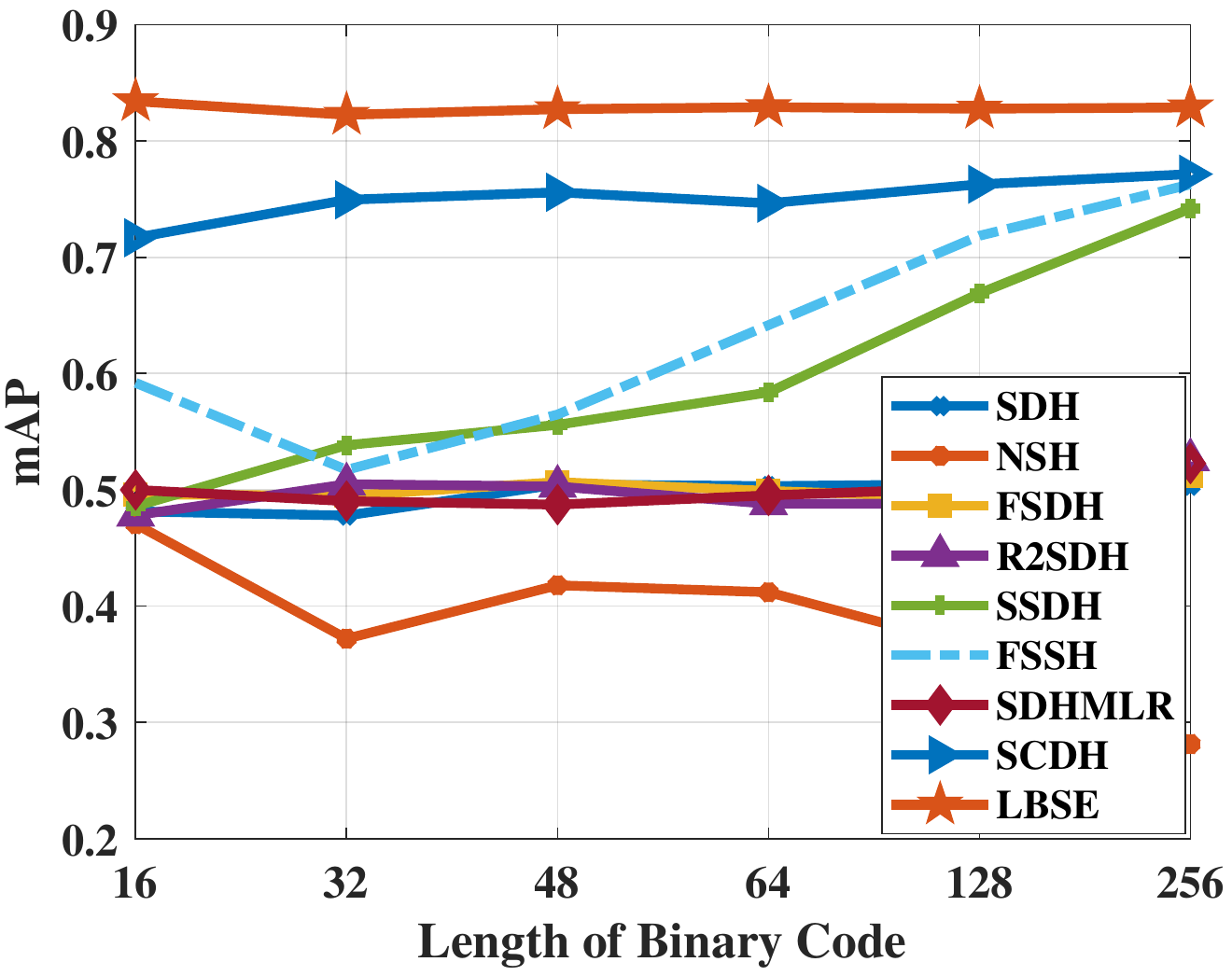}}
\caption{Performance in terms of the mAP scores with different lengths of binary code based on three benchmark datasets.}\label{mAPScore}
\end{figure}
\begin{figure}[htb]
\centering
\subfigure[BreaKHis 100X]{
\includegraphics[width=0.15\textwidth]{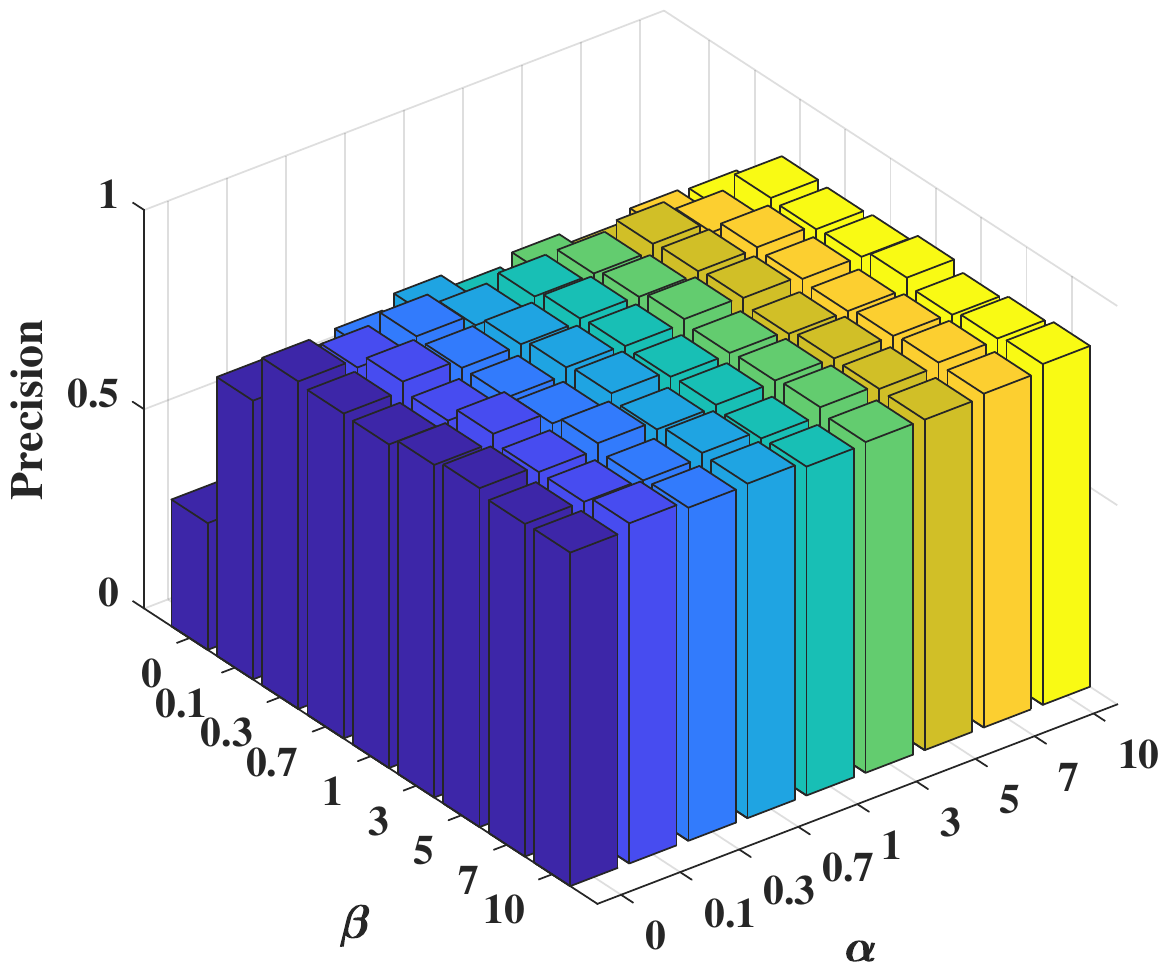}}
\subfigure[BreaKHis 200X]{
\includegraphics[width=0.15\textwidth]{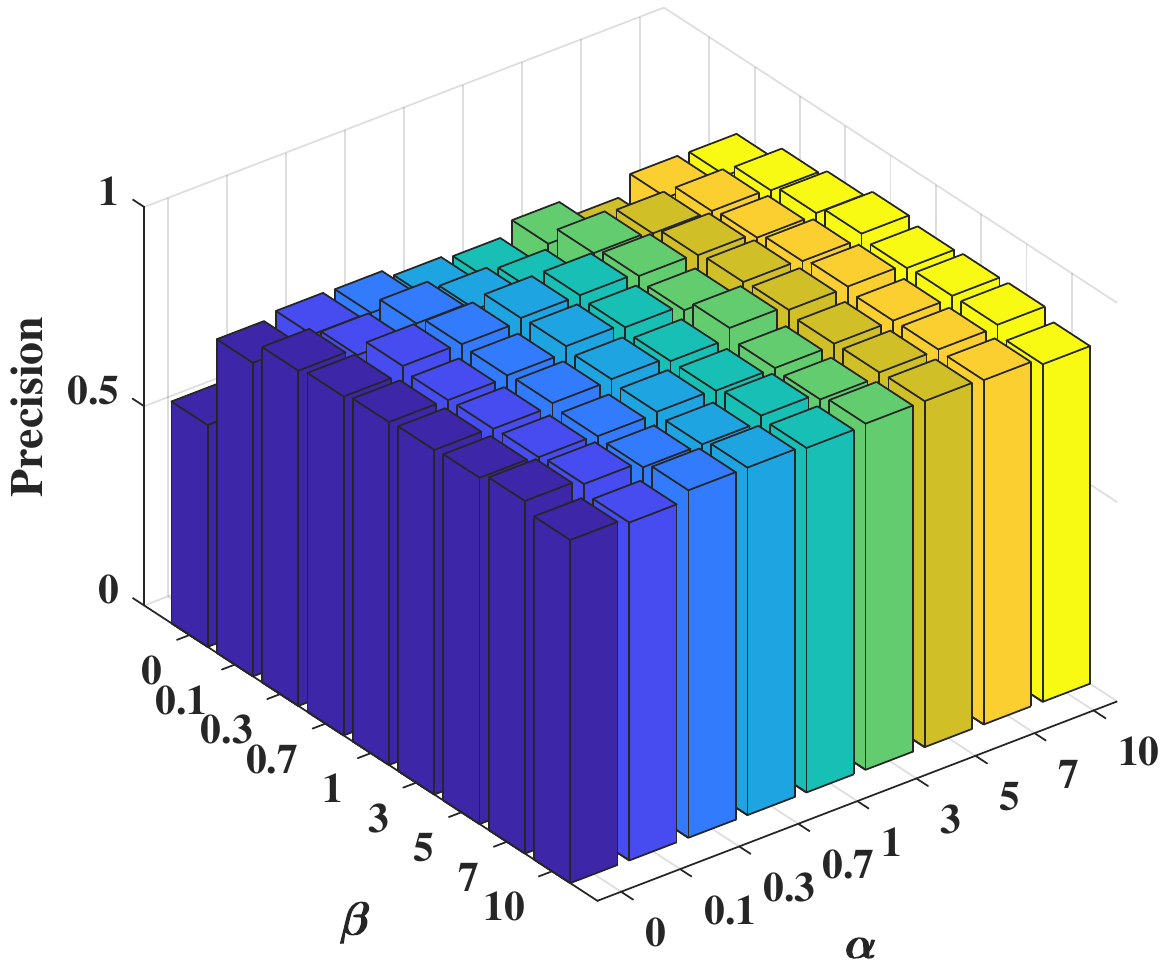}}
\subfigure[BreaKHis 400X]{
\includegraphics[width=0.15\textwidth]{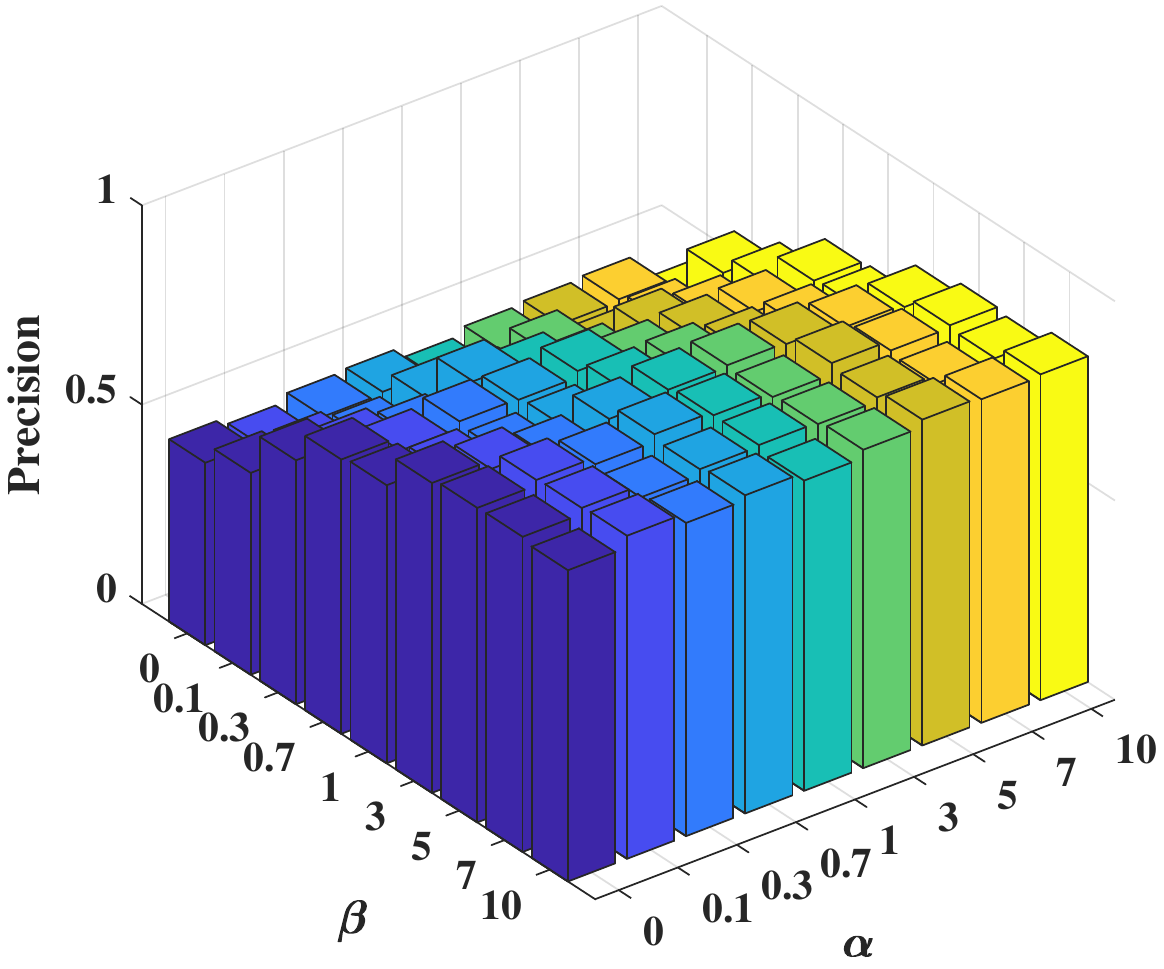}}
\caption{This figure shows the precision scores with different settings of $\alpha$ and $\gamma$ on three datasets when the length of binary code is 32.}\label{PPS}
\end{figure}
\subsection{Experimental Results and Analysis}

Table \ref{performance} shows the overall comparison of classification task on the three datasets. Limited by the space, we only present the classification results when the length of binary code is 32. It is worth nothing that the 32-bit embedding can represent $2^{32}$ categories, which may be large enough for classification task. And we can have found that these methods do not achieve much improvement when longer lengths of binary code are adopted. Furthermore, the proposed method shows satisfactory improvement as far as four evaluation matrices are concerned, indicating the proposed method perform well on classification task. The proposed method can outperform the other methods based on the BreaKHis 400X dataset distinctly. One possible reason is that the feature extracted from this dataset maybe not so differentiable, thus requiring a more superior classification model. The utilization of double supervision and discrete optimization makes the proposed method outperform in classification ability.

Fig. \ref{mAPScore}  exhibits the retrieval performance when the length of binary code ranging form \{ 16, 32, 48, 64, 128 and 256\}.  We can observe that the proposed method can acquire satisfactory performance under different scenarios of binary codes.  Moreover, the mAP scores achieved by the proposed method do not increase much as the length of binary code gets longer, indicating the proposed method can learn compact binary embedding and reduce the storage. In a word, the proposed method can obtain superior performance under different classification and retrieval tasks compared with the state-of-the-art hashing methods on the three  datasets.

We also conduct experiments with various parameter settings to verify the parameter sensitivity of the proposed LBSE. Due to limited space, we only showed the results about ${\alpha}$ and ${\gamma}$, which are most relevant to the performance of the proposed method. Fig. \ref{PPS}  shows the precision score of the LBSE, when ${\alpha}$ and ${\beta}$ are within a range; the proposed method exhibits acceptable stability and sensitivity.

\section{Conclusion}
  
In this paper, we propose a supervised  method for learning binary semantic embedding which focuses on preforming efficient classification and retrieval tasks and providing understanding auxiliary diagnosis based on histology images. In the proposed method, we consider double supervision,  bit uncorrelation and balance constraint, asymmetric  strategy and discrete optimization, while learning the effective and efficient binary semantic embedding, making it more suitable for precise classification and retrieval tasks.  The experimental results conducted on three benchmark datasets confirm the superiority of the proposed method under various scenarios.

\section{Acknowledgements}
This work was supported in part by the National Natural Science Foundation of China (61876098, 61671274, 61573219), National Key R$\&$D Program of China (2018YFC0830100, 2018YFC0830102) and special funds for distinguished professors of Shandong Jianzhu University.

\bibliographystyle{IEEEbib}
\bibliography{refs}
\end{document}